\documentclass[letterpaper, 10 pt, conference]{ieeeconf}  
\usepackage{graphicx}
\usepackage{cite}
\usepackage{amsmath,amssymb,amsfonts}

\usepackage{array}
\usepackage{subcaption,graphicx}
\usepackage{color}
\usepackage{bm}
\usepackage{cite}
\usepackage{stfloats}
\usepackage{subcaption}
\usepackage{caption}
\usepackage{url}

\IEEEoverridecommandlockouts                              
\title{\LARGE \bf
An Optimal LiDAR Configuration Approach for Self-Driving Cars
}

\author{Shenyu Mou$^{1}$, Yan Chang$^{2}$, Wenshuo Wang$^{3}$, and Ding Zhao$^{4}$
\thanks{This study is supported by Denso International America, Inc.}
\thanks{S. Mou is with the Department of Electrical Engineering, University of Michigan, Ann Arbor, MI, 48105.
        {\tt\small mshenyu@umich.edu.}}%
\thanks{Y. Chang, W. Wang, and D. Zhao are with the Department of Mechanical Engineering, University of Michigan, Ann Arbor, MI, 48105.
        {\tt\small yanchang@umich.edu; wwsbit@gmail.com; zhaoding@umich.edu.}}%
}

\begin{document}

\maketitle
\thispagestyle{empty}
\pagestyle{empty}

\begin{abstract}
LiDARs plays an important role in self-driving cars and its configuration such as the location placement for each LiDAR can influence object detection performance. This paper aims to investigate an optimal configuration that maximizes the utility of on-hand LiDARs. First, a perception model of LiDAR is built based on its physical attributes. Then a generalized optimization model is developed to find the optimal configuration, including the pitch angle, roll angle, and position of LiDARs. In order to fix the optimization issue with off-the-shelf solvers, we proposed a lattice-based approach by segmenting the LiDAR's range of interest into finite subspaces, thus turning the optimal configuration into a nonlinear optimization problem. A cylinder-based method is also proposed to approximate the objective function, thereby making the nonlinear optimization problem solvable. A series of simulations are conducted to validate our proposed method. This proposed approach to optimal LiDAR configuration can provide a guideline to researchers to maximize the utility of LiDARs.

\end{abstract}

\section{Introduction}

LiDAR has been widely used in self-driving cars because its powerful capabilities to gather the profile-related information of surroundings and help read depth information, which is far beyond the ability of cameras. The application of LiDAR in autonomous vehicle system has existed for a long time. Wijesoma and his teammates has finished a research where laser has been used for road boundary detection\cite{road-boundary}. MacLachlan used a scanning laser rangefinder to detect objects and their future path\cite{object-detection-prediction}. The reasons that LiDAR can be so helpful for autonomous vehicle are that LiDAR is highly precise and the point clouds from LiDAR offer rich information of the environment which can be used to achieve complex missions. Zhang used velodyne point cloud to estimate the ground truth of autonomous car's path\cite{zhang2014loam} and achieved the top one in the competition of KITTI Odometry by taking advantages of LiDAR point clouds.

\begin{figure}
    \centering
    \begin{subfigure}[b]{.45\linewidth} \includegraphics[width=\linewidth]{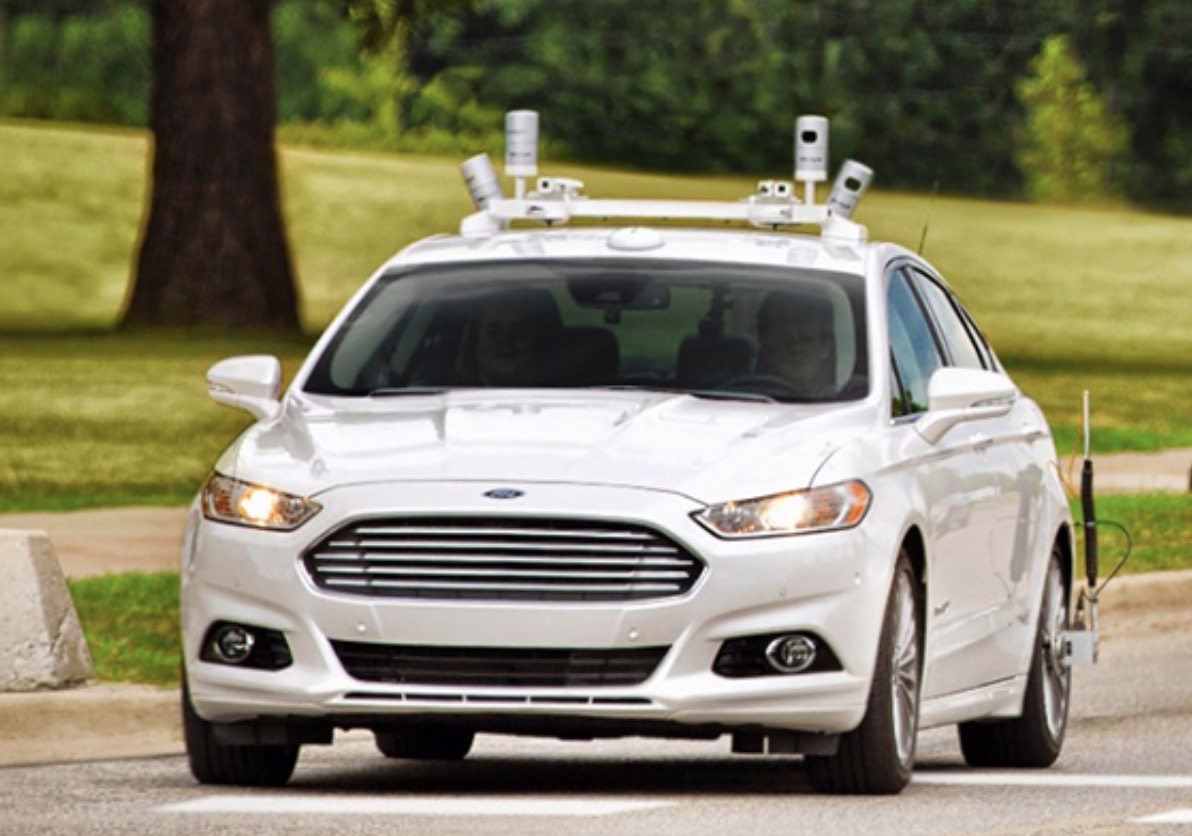}
        \caption{Ford}
    \end{subfigure}%
        \begin{subfigure}[b]{.45\linewidth} \includegraphics[width=\linewidth]{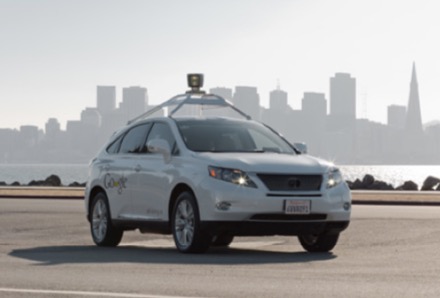}
        \caption{Waymo}
    \end{subfigure}%
    
    \begin{subfigure}[b]{.45\linewidth} \includegraphics[width=\linewidth,height=0.7in]{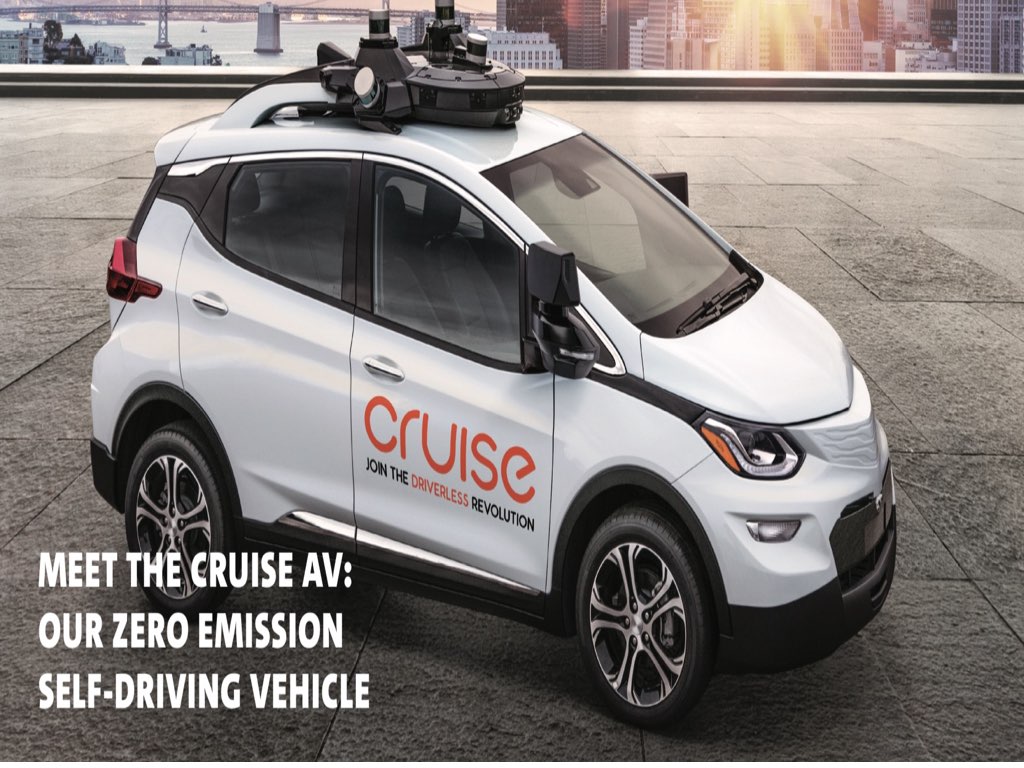}
        \caption{Cruise}
    \end{subfigure}%
    \begin{subfigure}[b]{.45\linewidth} \includegraphics[width=\linewidth]{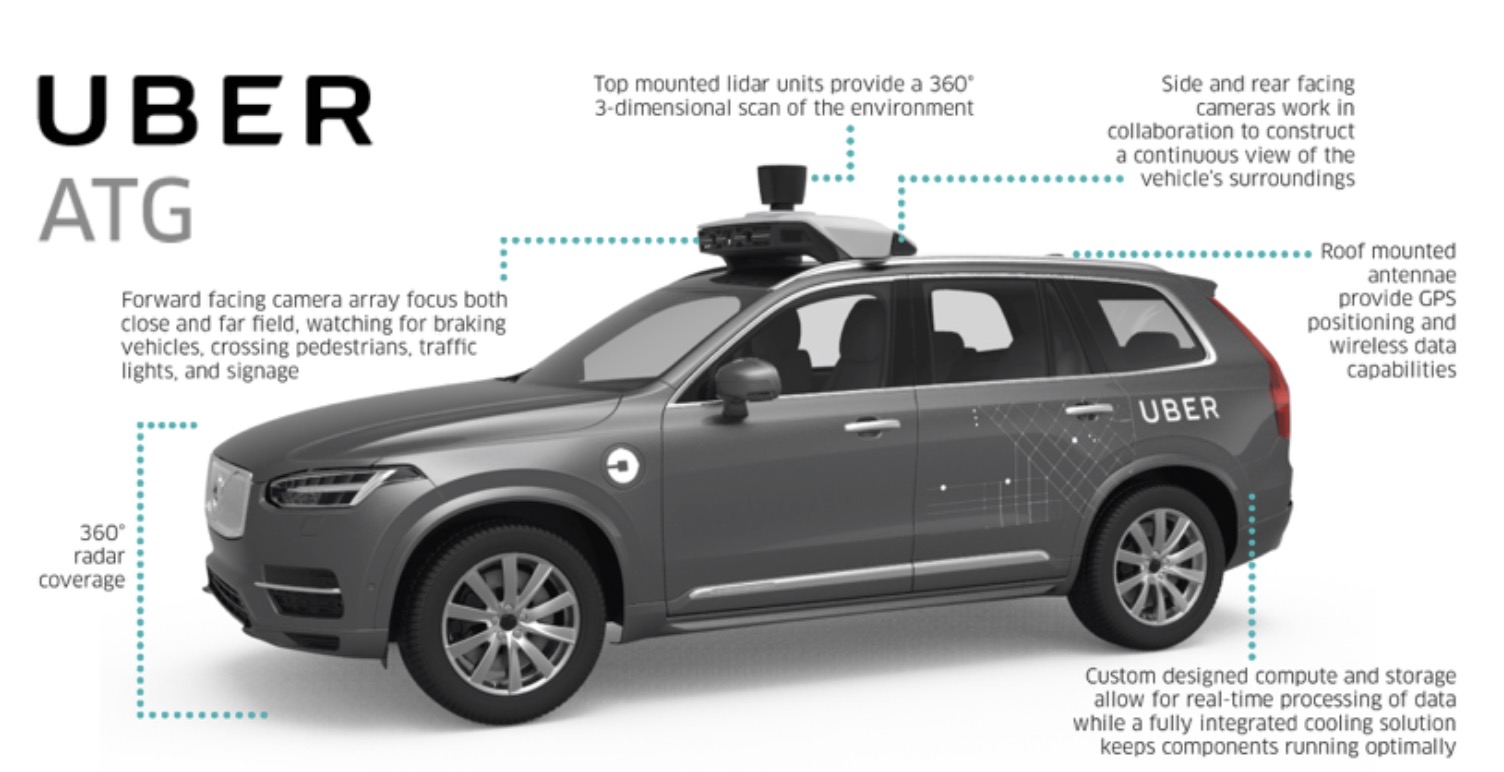}
        \caption{Uber}
    \end{subfigure}%
    
    \begin{subfigure}[b]{.45\linewidth} \includegraphics[width=\linewidth]{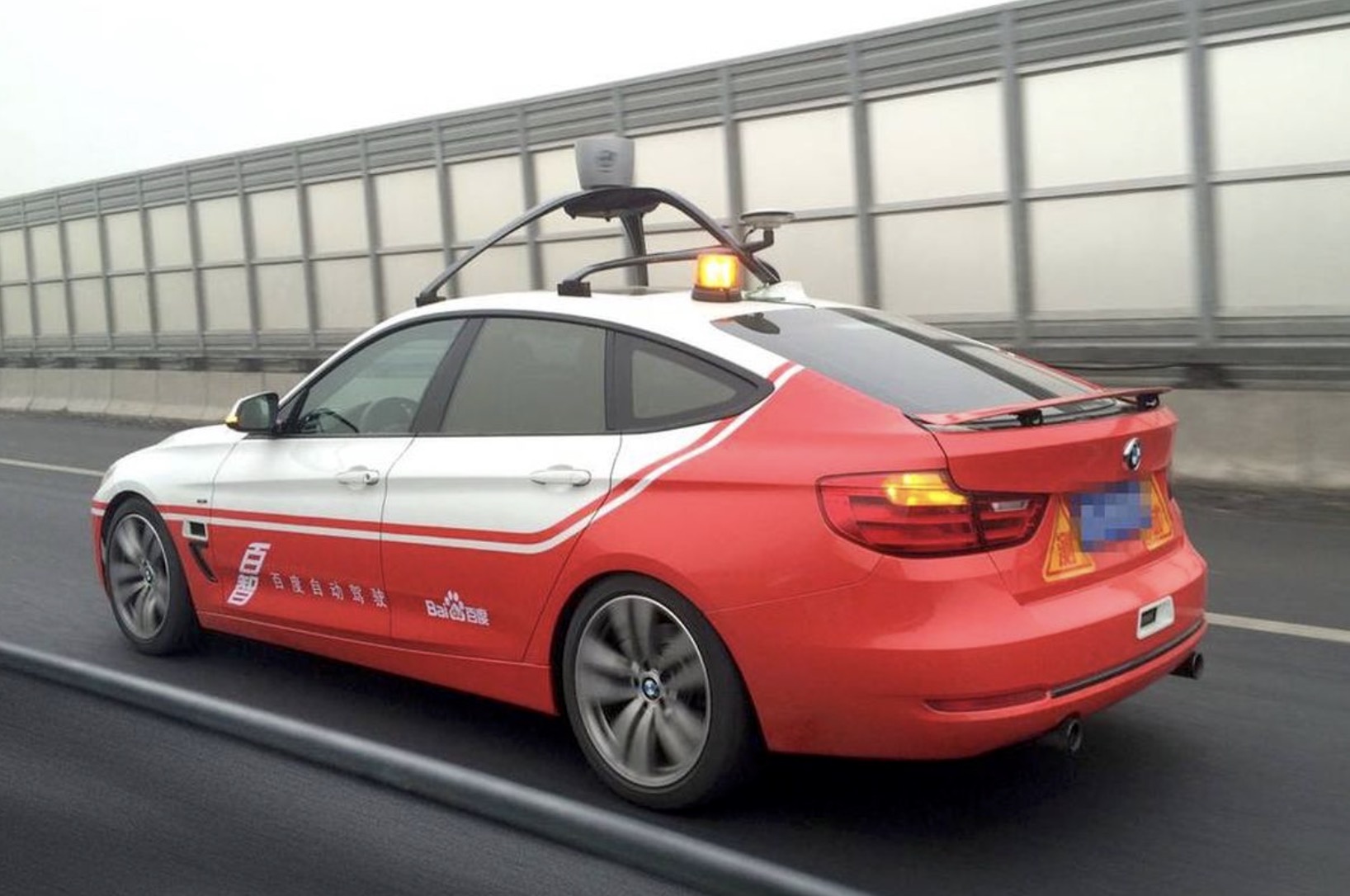}
        \caption{Baidu}
    \end{subfigure}%
        \begin{subfigure}[b]{.45\linewidth} \includegraphics[width=\linewidth]{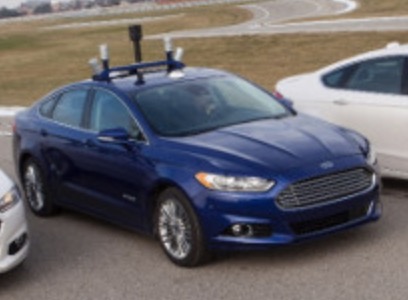}
        \caption{University of Michigan}
    \end{subfigure}%
    
    \begin{subfigure}[b]{.45\linewidth} \includegraphics[width=\linewidth]{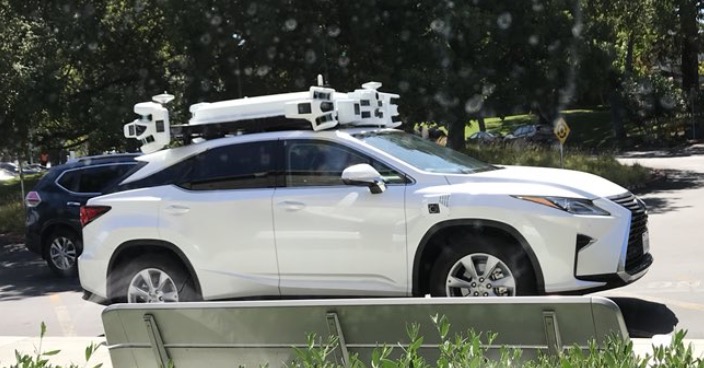}
        \caption{Apple}
    \end{subfigure}%
    \begin{subfigure}[b]{.45\linewidth} \includegraphics[width=\linewidth]{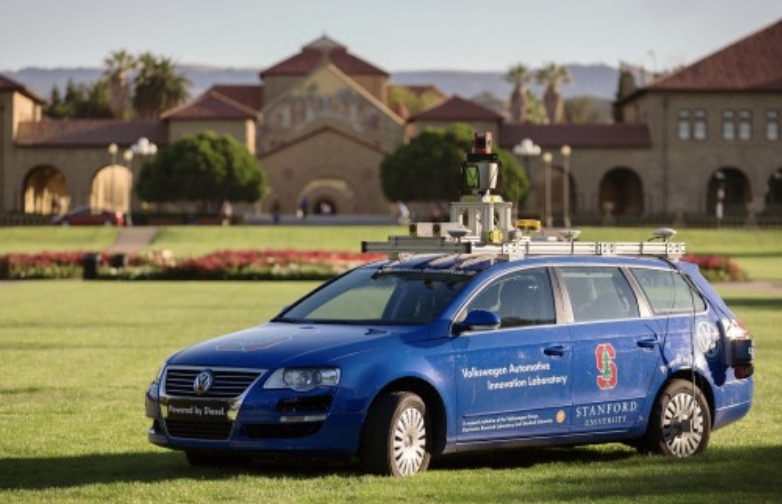}
        \caption{Stanford University}
    \end{subfigure}%
    \caption{Examples of the LiDAR configurations with different existing autonomous cars, from left to right and top to bottom: Ford\cite{fordweb}, Cruise\cite{gmweb}, Waymo\cite{waymoweb}, Uber\cite{uberweb}, Baidu\cite{baiduweb}, Apple\cite{apple}, UM\cite{umweb}, and Standford\cite{stanfordweb}.}
\label{fig:example_configuration}
\end{figure}

Recently, autonomous vehicle suppliers and self-driving research centers deploy LiDARs with various configurations. For instance, Ford equips their autonomous cars with four velodyne-16 LiDARs, two at each sides on the roof of an autonomous car with an roll angle between them. Waymo installs one velodyne-64 LiDAR on the roof for perception. Some other famous self-driving solution suppliers hold other LiDAR configuration strategies as shown in Fig. \ref{fig:example_configuration} and Table \ref{configlist}. However, it is still not fully clear which strategy is the best one and how many LiDARs is the best selection. In general, more LiDARs can provide more precise information to autonomous vehicles, but lead to information redundancy and high costs. Therefore, a trade-off between high-resolution information and low-cost of LiDARs should be balanced. An optimal configuration should at least follow the two abilities -- highly informative and low cost -- that is, it is capable of capturing surrounding information as much as possible at low cost of LiDARs. Therefore, an optimal solution to LiDAR configuration is desired for a given task.

\begin{table*}[t]
  \begin{center}
    \caption{$360^\circ$ LiDAR Configurations of Different Self-Driving Teams}
    \begin{tabular}{l l l l}
    \hline \hline
      \textbf{Team} & \textbf{LiDAR Type} & \textbf{Number} & \textbf{Layout} \\
      \hline
      Ford & Velodyne-16 & 4 & 2 at each side on top roof \\
      Cruise & Velodyne-16 & 5 & 2 at each side with 1 at the middle front on top roof\\
      Waymo & Velodyne-64 & 1 & 1 at top center on the roof \\
      Uber & Velodyne-64 & 1 & 1 at top center on the roof \\
      Baidu & Velodyne-16/64 & 2/1 & 1 Velodyne-64 at top with 1 Velodyne-16 at each side\\
      Apple & Velodyne-16 & 12 & 6 at front and 6 at rear on the roof\\
      UM perl lab & Velodyne-16/64 & 4/1 & 2 Velodyne-16 at each side; 1 Velodyne-64 on top center \\
      Stanford Driving Team & Velodyne-64 & 1 & 1 at top center on the roof \\
      \hline  \hline
    \end{tabular}
    \label{configlist}
  \end{center}
\end{table*}

Some existing literature dedicates their efforts on optimal 2D sensor configuration \cite{Zhang}. Optimal solution on 3D perception sensor configuration has also been discussed. For instance, Dybedal and Hovland introduced the generally 3D optimal camera configuration \cite{Dybedal}. The authors proposed an optimal 3D camera layout method to acquire the largest field of view through the optimal camera configuration. The space was segmented into finite small cubes and the optimal 3D camera configuration was evaluated through the number of cubes in the observation range of cameras. Banta and Abidi described a system that determines the optimized range sensor positions for reconstruction \cite{Banta}. Rahimian and Kearney applied optimal camera layouts for motion capture systems \cite{Rahimian}. In the aforementioned research, the environment information collected by range sensors were considered as {\it continuous} by fully exploiting the property of cameras. However, it is great challenging to determine an optimal LiDAR configuration using such approaches since LiDARs have spatial and {\it discrete} features, which greatly differs from cameras. LiDAR usually suffers a perception with blind areas growing along the perception distance, therefore the sparsity should be considered as a factor when figuring out the optimal solution of LiDAR configuration. To our the best of knowledge, rare publications about optimal sensor configuration focus on LiDAR layout or configuration for self-driving cars, and there not exists a generalized solution to this problem.

Toward this end, we systematically investigate this problem in this paper. Our main contributions are threefold: 1) we raise the LiDAR configuration problem in self-driving cars for the first time to our the best knowledge; 2) A generalized optimization model is developed for 3D LiDAR configuration with concerning sparsity of LiDAR perception; 3) A lattice-based approach combining with a cylinder-based approximation is developed to make optimization problem solvable. 

The organization of this paper is as follows. Section II details the LiDAR configuration structure. Section III introduces the optimal problem formulation of LiDAR configuration. Section IV shows the simulation results and analysis. Finally, the conclusion and future work are discussed in Section V.

\section{LiDAR Configuration Structure}

The goal of the optimal LiDAR configuration is to acquire as much information as of surroundings under the limited number of LiDARs provided. Hence in order to achieve this goal, we should define the most informative perception.

\begin{figure}[t]
\centering
\includegraphics[width=0.9\linewidth]{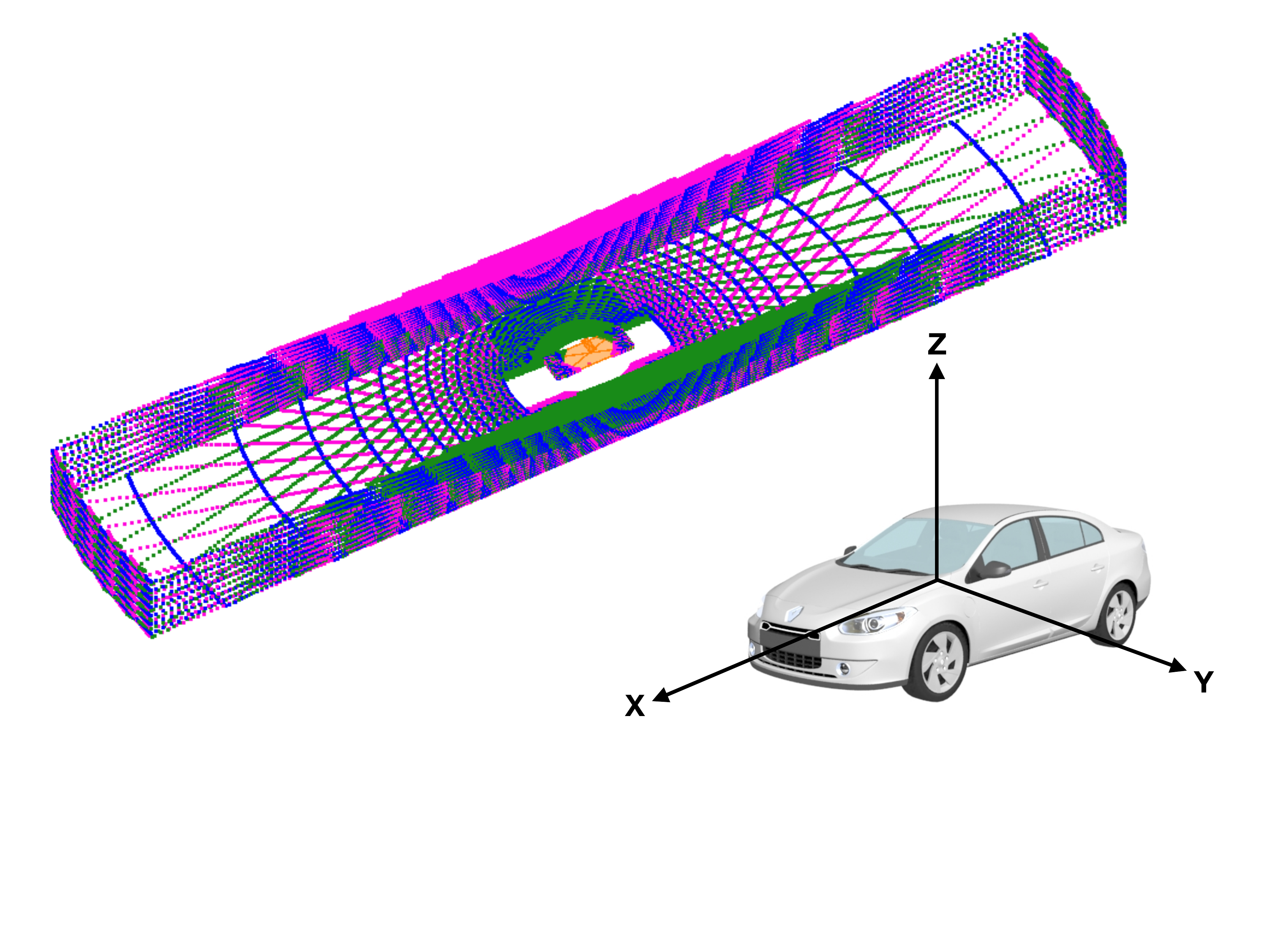}
\caption{Range of interest (ROI) of autonomous cars carrying three LiDARs.}
\label{fig:ROI}
\end{figure}

\subsection{Range of Interest}
The surroundings far away from the self-driving car will have less influence on it. In order to simplify the configuration problem, here we only concern LiDAR's range of interest (ROI) into a truncated space where we intend to acquire as much information as possible, as shown in Fig. \ref{fig:ROI}. The space outside the ROI are not considered. The ROI can be defined as a cube with preset conditions and the origin of $x-y$ planet of ROI is aligned with the origin of $x-y$ planet of self-driving car. For example, the ROI can be defined with height of 5 m, width of 9 m, and length of 80 m.

\subsection{Informative Perception}

\begin{figure}[t]
\centering
\includegraphics[width=0.4\textwidth]{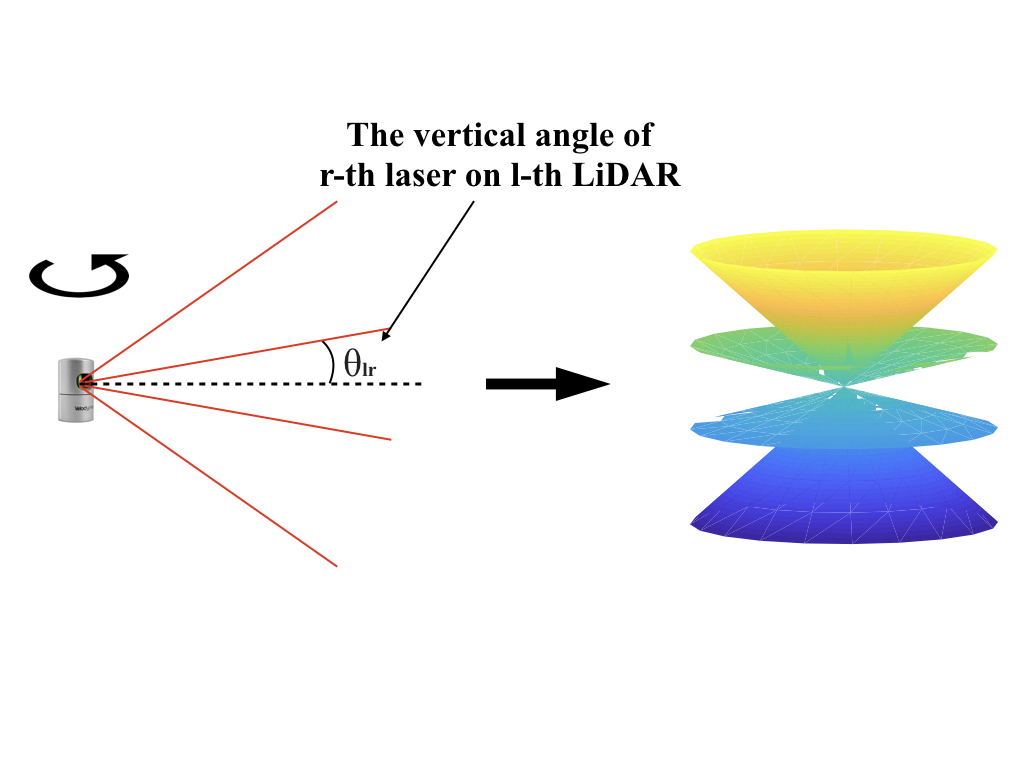}
\caption{A LiDAR with lasers forming cones by a $360$-degree rotation.}
\label{fig:cone}
\end{figure}

The perception data is collected from the trajectory of lasers generated by LiDAR, featured by range and bearing. Therefore, the perception of LiDARs is sparse and {\it discrete} in a 3D space over several routes of lasers. With the fact that the horizontal resolution of LiDARs is constant and can not get improved by LiDAR configurations, we assume the horizontal resolution of LiDARs is 0. Namely, when a LiDAR takes a $360$-degree rotation, the trajectory of each laser on this LiDAR will form a cone, as shown in Fig. \ref{fig:cone}, where $\theta_{lr}$ is the beam angle. Therefore, LiDAR can be modeled as several cones sharing one  vertical axis. With all the cones formed by all LiDARs, ROI can be segmented into subspaces bounded by surfaces of these cones and ROI boundaries.

To achieve the most informative perception, the laser should be able to detect as small as possible objects in the ROI. Therefore, each subspace should be small enough with limited number of LiDARs. Due to the diversity in spatial shapes of the segmented subspaces, here we introduce the inscribed sphere of each subspace to describe their size. The most informative perception of LiDARs system is referred as the case when the radius of the largest inscribed sphere is minimized. In this way, the LiDAR optimization configuration is transformed into a minmax optimal problem.

\subsection{Subspace Segmentation}

\begin{figure}[t]
    \centering
    \begin{subfigure}[b]{.4\linewidth} \includegraphics[width=\linewidth]{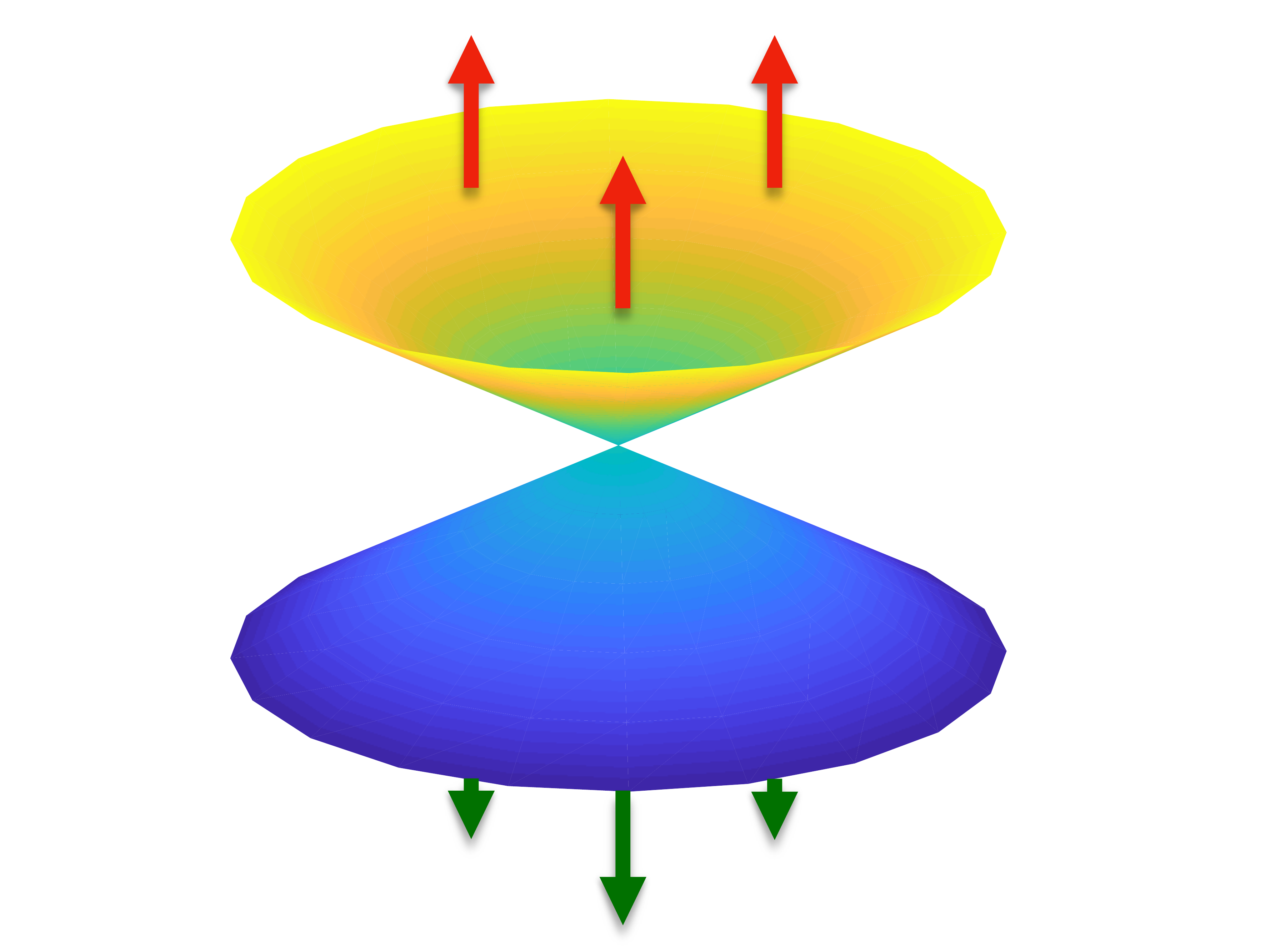}
        \caption{Empty subspace}
    \end{subfigure}%
    \begin{subfigure}[b]{.4\linewidth} \includegraphics[width=\linewidth]{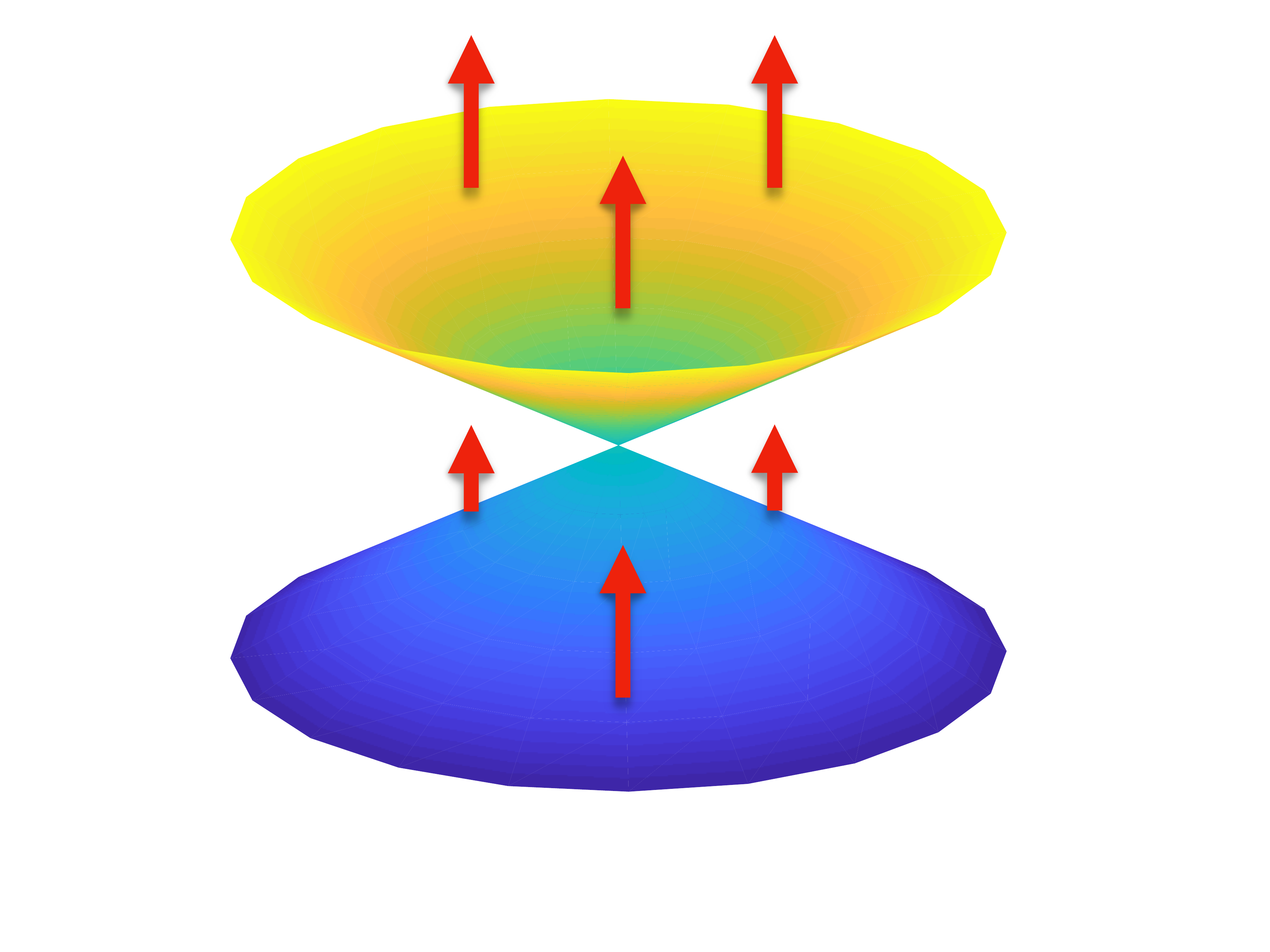}
        \caption{Nonempty subspace}
    \end{subfigure}%
    \caption{Left: An empty subspace occurs when there is no intersection of two-side spaces selected from two cones. Right: A nonempty subspace occurs when there is an intersection of two-side spaces selected from two cones.}
    \label{fig:empty_nonempty}
\end{figure}

Each cone in Fig. \ref{fig:cone} segments the ROI into two subspaces: upward side and downward side. Thus, $N_l$ LiDARs with $N_r$ lasers segment the ROI into $2^{N_l\times N_r}$ subspaces and many of subspaces are empty due to the physical constraints of LiDARs, as shown in Fig. \ref{fig:empty_nonempty}. For example, a subspace will be empty when the upward side of a upper laser with a larger beam angle combines with the downward side of a lower laser with a smaller beam angle.

\begin{figure}[t]
\centering
\includegraphics[width=0.4\textwidth]{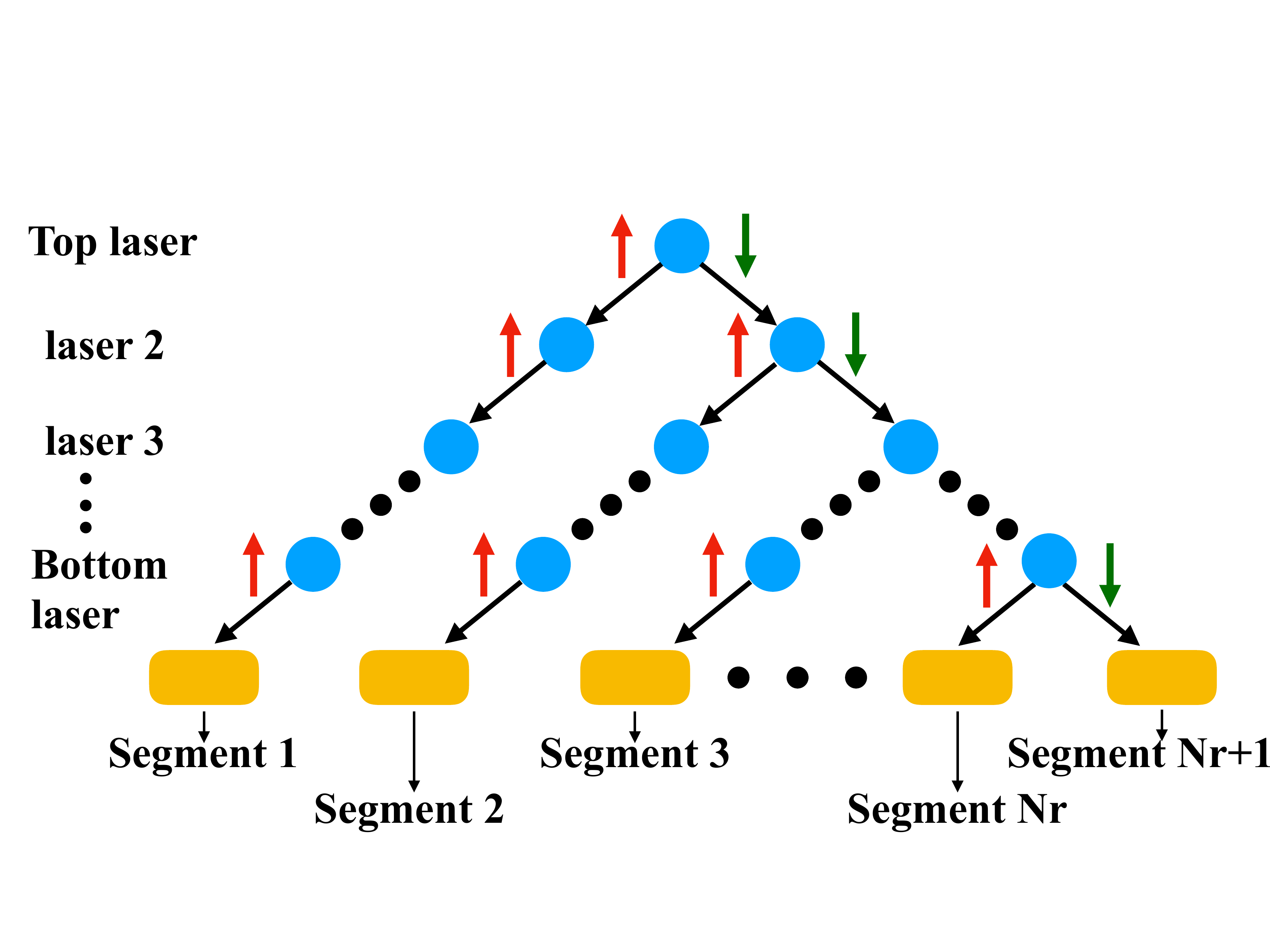}
\caption{Combination rule: segmenting the ROI according to combination of side selection of cones for one LiDAR.}
\label{fig:tree}
\end{figure}

To potentially remove these empty subspaces, we follow the physical constraints of LiDAR and design a combination rule to segment the ROI, described as a tree (Fig. \ref{fig:tree}). Different paths represent different combination results. From each path, the binary flag $f_{lr}$ to change the sign of inequality for $lr$-th cone can be found. Thus, one LiDAR with $N_r$ lasers can segment the ROI into $N_r + 1$ nonempty subspaces and $N_l$ LiDARs can generate $(N_r + 1)^{N_l}$ nonempty subspaces at most.

\subsection{Radius of Inscribed Sphere of Subspaces}
Subspaces bounded by cones and ROI boundaries are featured with irregular shapes. To directly find the inscribed sphere of any irregular shapes using a analytical way is difficult in our optimization case due to the lack of analysis formula to express it. Moreover, the difficulty is promoted by the need to solve a sub-optimal problem with decision variables of free $[x,y,z]$ coordinates that are independent with the configuration variables of LiDARs. Instead, we propose an approximate representation of the inscribed sphere as follows. 

Considering a small instance in 2D case, a subspace can be generated by the combination of side selection of lines. Suppose there are 3 lines:
\begin{align}
l_1 : y = k_1x+b_1 \\
l_2 : y = k_2x+b_2 \\
l_3 : y = k_3x+b_3
\end{align}
where the configuration of lines is featured by $[k_1,b_1,k_2,b_2,k_3,b_3]$. Without using the analytically explicit formula to represent the radius of the inscribed sphere, we turn to solve an optimization problem with variables as free $[x,y]$ instead of $[k_1,b_1,k_2,b_2,k_3,b_3]$. Thus, this sub-optimization problem is formulated by

\begin{equation}
\max \ \min[d_1(x,y),d_2(x,y),d_3(x,y)]
\end{equation}
where $[d_1(x,y),d_2(x,y),d_3(x,y)]$ are the distances from a point in the triangle formed by $l_1$, $l_2$ and $l_3$ to the three lines, respectively. We then solve an optimization problem with decision variables of $[k_1,b_1,k_2,b_2,k_3,b_3]$ which are independent with the decision variables of $[x,y]$ of the sub-optimization problem. However, the whole optimization problem can be much complex and hard to solve.

\begin{figure}[h]
\centering
\includegraphics[width=0.4\textwidth]{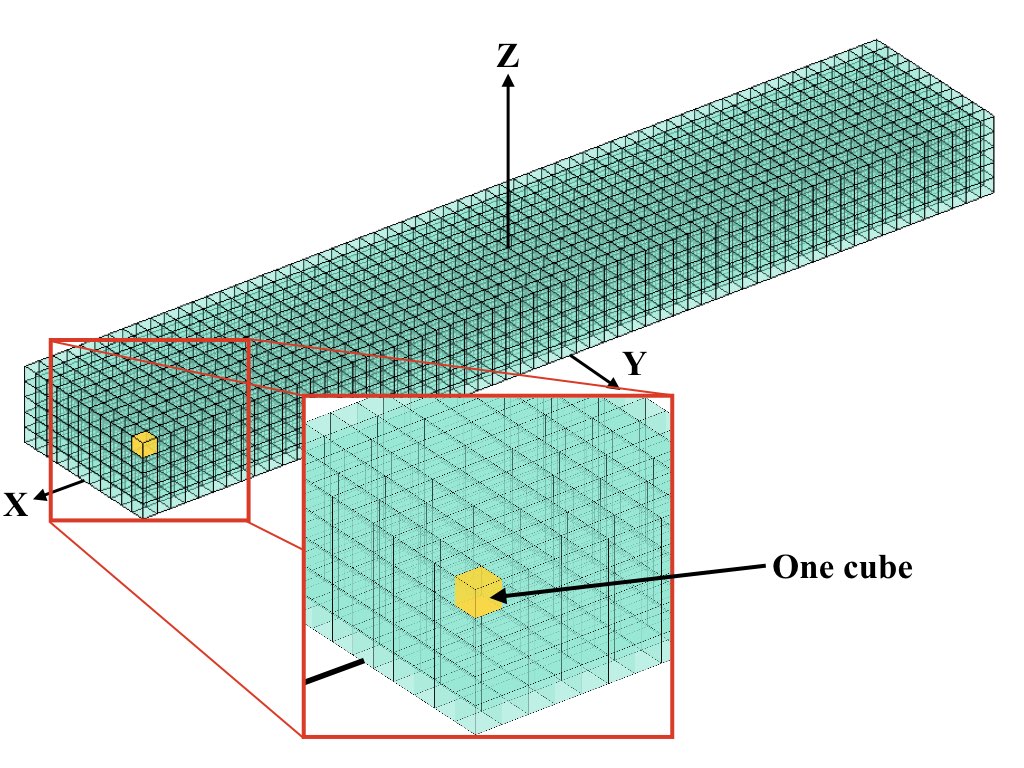}
\caption{Discretization of ROI by cubes.}
\label{cubes}
\end{figure}

To avoid a sub-optimization problem, we make the decision variables $[x,y,z]$ of the sub-optimization problem constant by discretizing them with cubes Fig. \ref{cubes}, inspired by \cite{Dybedal}. In this way, the decision variables of the sub-optimization problem turn to be many constant discrete numbers. As a result, the radius of the inscribed sphere is represented by the number of cubes whose details will be introduced in Section III-B. And the payment is the extended number of constraints and new decision variables for each cube.

\begin{figure}[h]
\centering
\includegraphics[width=0.4\textwidth]{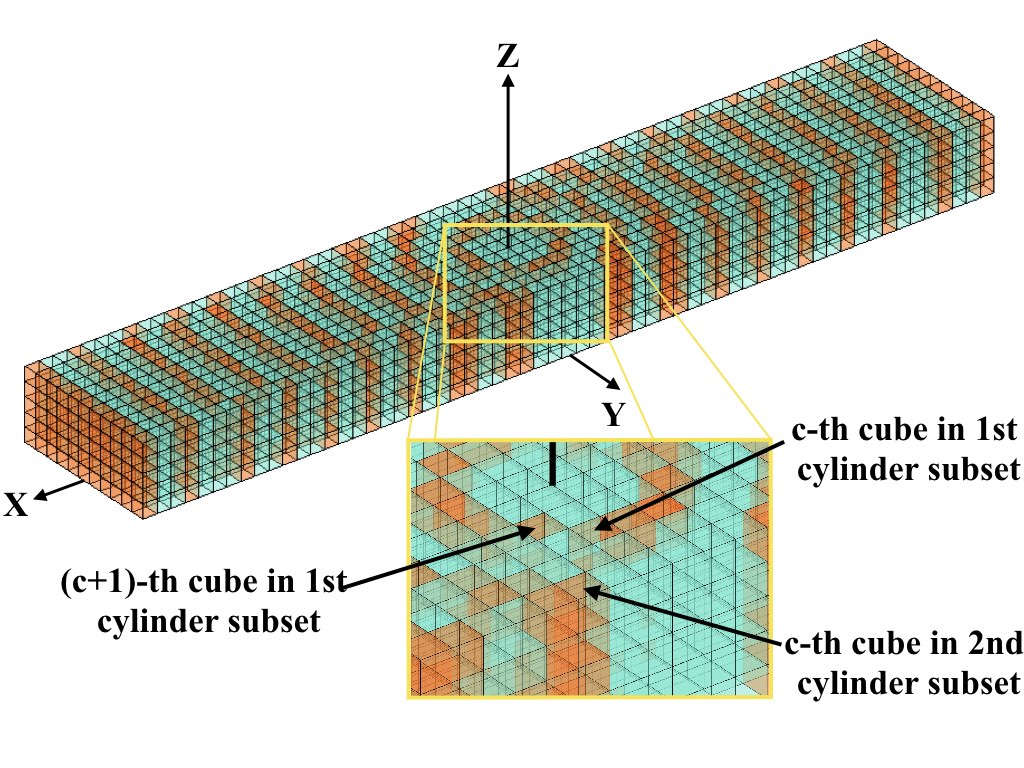}
\caption{Select cubes in the ROI with the help of concentric cylinders.}
\label{cylinder}
\end{figure}

\begin{figure}[h]
\centering
\includegraphics[width=0.4\textwidth]{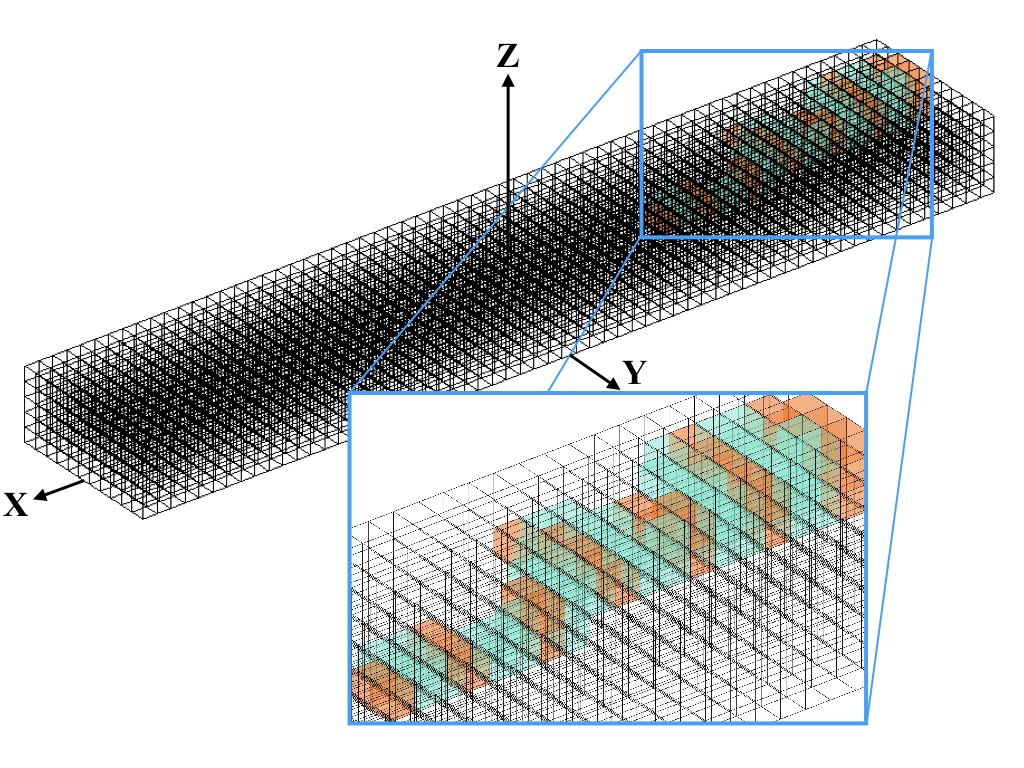}
\caption{Choose the subset that contains most cubes to represent the radius of inscribed sphere for one subspace.}
\label{subspace}
\end{figure}

To find the approximation of inscribed sphere radius of a subspace, we need to select a subset, from all the cubes in a subspace, that can represent the inscribed sphere radius. Here we employ concentric cylinders with their center fixed with the origin of self-driving car coordinate system, as shown in Fig. \ref{cylinder}. The radius gaps between two cylinders are parameters that can be tuned. For each subspace, its cubes that intersect with the side surfaces of different cylinders are selected as different subsets. The subset that contains maximum number of cubes is selected to represent the radius of the inscribed sphere in this subspace as shown in Fig. \ref{subspace}.

\section{Problem Formulation}

\subsection{Cone Representation}

In order to determine whether a cube center is at the upward side of a cone or downward side of a cone, a convenient way is to transfer the cube center into the local coordinate system of a cone, which is defined as the following:
\begin{enumerate}
\item The origin is on the start point of the laser that forms the cone.
\item The $x,y,z$ axises are transformed by the configuration parameter of the LiDAR. 
\end{enumerate}

In the local coordinate system of LiDARs, the cones formed by the lasers on this LiDAR can be represented as

\begin{equation}
z^L = \tan\theta_{lr}\sqrt{(x^L)^2+(y^L)^2}
\end{equation}
where $x^L$,$y^L$, and $z^L$ are the local coordinates of a LiDAR. If a point is at upward side of this cone, its local coordinates satisfy the constraint

\begin{equation}
z^L - \tan\theta_{lr}\sqrt{(x^L)^2+(y^L)^2} > 0
\end{equation}
and otherwise, its local coordinates satisfy the constraint 

\begin{equation}
z^L - \tan\theta_{lr}\sqrt{(x^L)^2+(y^L)^2} < 0.
\end{equation}

\subsection{Inscribed Sphere Representation}

A subspace is a space where all the points in it satisfy the side constraints discussed in Section III-A, following the combination rule discussed in Section II-C shown in Fig. \ref{fig:tree}. To deal with one optimization problem with another sub-optimization problem holding independent decision variables, we first represent a subspace by all cubes shown in Fig. \ref{cubes}. The centers of these cubes are tested by the side constraints discussed in Section III-A, following the combination rule discussed in Section II-C. And the flag $f_{lr}$ for selecting the side of the $lr$-th laser is determined according to the combination rule. And then we represent the radius of inscribed sphere of a subspace by the cylinders introduced in Section II-D.

The logic to determine whether a cube is in some subspace is: if the cube's center fails one of the subspace boundary constraints then it is not in the subspace, otherwise it is in the subspace. To mathematically describe the subspace representation, we formulate the constraints as the following.

For the $c$-th cube in the subset of the $k$-th cylinder, with center at $[x_c,y_c,z_c]$ in local LiDAR coordinates, we have


\textbf{For} $s$ = 1 \textbf{to} $N_{ss}$

\ \ \textbf{For} $l$ = 1 \textbf{to} $N_l$

\ \ \ \ \textbf{For} $r$ = 1 \textbf{to} $N_r$

$$
\begin{bmatrix}
	x_c^L\\
    y_c^L\\
	z_c^L\\
    1
\end{bmatrix}
= \mathbf{H_{\textit{l}}}
\begin{bmatrix}
	x_c\\
    y_c\\
	z_c\\
    1
\end{bmatrix}
$$

\ \ \ \ \ \ \textbf{IF}
$$
 	(z_c^L - \tan\theta_{lr}\sqrt{(x_c^L)^2+(y_c^L)^2})f_{lr} >0
$$

\ \ \ \ \ \ \textbf{THEN}
$$
	E_{sck}=0
$$

\ \ \ \ \ \ \textbf{ELSE}
$$
	E_{sck}=1
$$

\ \ \ \ \textbf{EndFor}

\ \ \textbf{EndFor}

\textbf{EndFor}\\
where $\mathbf{H_{\textit{l}}}$ is the transformation matrix to transfer the global cube coordinates into the local ones of $l$-th LiDAR derived in appendix A, and E$_{sck}$ is the binary value, determining if the $c$-th cube in the subset of the $k$-th cylinder is inside the $s$-th subspace. And the logic constraints are introduced in part B of appendix. As a result, the representation of the inscribed sphere radius of the $s$-th subspace (Fig. \ref{subspace}) can be 

\begin{equation*}
F_s(\textbf{C}) =\max\limits_{k=1}^{N_k} \sum\limits_{c=1}^{N_c}E_{sck}
\end{equation*}

With all the constraints generated, the decision variables of configuration of LiDARs $\mathbf{C} = [\boldsymbol{X},\bm{Y},\bm{Z},\bm\beta,\bm\gamma]$ can be optimized by the object function

\begin{equation*}
\overline{\mathbf{C}} = \mathop{\arg\min}_{\mathbf{C}}\max\limits_{s=1}^{N_ss} F_s(\textbf{C})
\end{equation*}
which solves the minimal value of the largest radius among all the inscribed sphere radii. And the results of solved configuration are the optimal configuration ensuring the most informative perception of LiDARs. Because the object function minimizes a max value of a set of max values, it can change to a more general format

\begin{equation*}
\overline{\mathbf{C}} = \mathop{\arg\min}_{\mathbf{C}}\max\limits_{s=1,k=1}^{N_ss,N_k} E_{sck}
\end{equation*}

\section{Simulation Results and Analysis}
In this section, we will show the effectiveness of our proposed approach to investigate the optimal LiDAR configuration by giving a specific case study.
\subsection{Simulation Settings}
This case is simulated in typical laptop with 3.5 GHz Intel Core i7 processor, 8 GB 2133 MHz LPDDR3 memory. The running time is about 1 hours. The optimization solver used is Gurobi (version 7.5.2) due to its capability of solving mix integer programming and its compatibility with many languages such as matlab, python, C and C++. For our case, matlab and C serves as the interfaces of Gurobi.

\subsection{Simulation Results}
For the sake of computational burden, the case we will show only considers 2 LiDARs ($N_r=2$) and each LiDAR has 2 lasers ($N_l=2$) with pitch angles as $\pm 10^\circ$ as shown in Table \ref{table:parameters}.

\begin{table}[h]
  \centering
    \caption{Case parameters}\label{table:parameters}
    \begin{tabular}{ll} 
    \hline\hline
      Variable & Value\\
      \hline
      $\theta_{l1}$ & $10^\circ$\\
      $\theta_{l2}$ & $-10^\circ$\\
      $x$ range of ROI & $[-8.5, 8.5]$ m\\
      $y$ range of ROI & $[-2.5, 2.5]$ m\\
      $z$ range of ROI & $[0,5]$ m\\
      \hline\hline
    \end{tabular}
\end{table}

In these case, to linearize the constraints, we treat the cones formed by lasers as pyramids with 4 side surfaces, which generates 4 binary variables judging whether a cube satisfies the constraints of the 4 side surfaces and 1 binary variable $d_la$ that is \textbf{\textit{OR}} logic of these 4 binary variables. Also, there is another binary variable $d_seg$ that is the \textbf{\textit{AND}} logic of all $d_la$ of different lasers. Then a binary variable $d_c$ is generated to be the \textbf{\textit{AND}} logic of all $d_seg$ of different LiDARs to judge whether a cube is in a subspace. Next, a binary variable $d_ss$ counts the sum of $d_c$ to represent the number of cubes in one subspace. Thus all $d_ss$ finally generate 24307 variables and 48618 constraints.

\begin{figure}[t]
\centering
\includegraphics[width=0.5\textwidth]{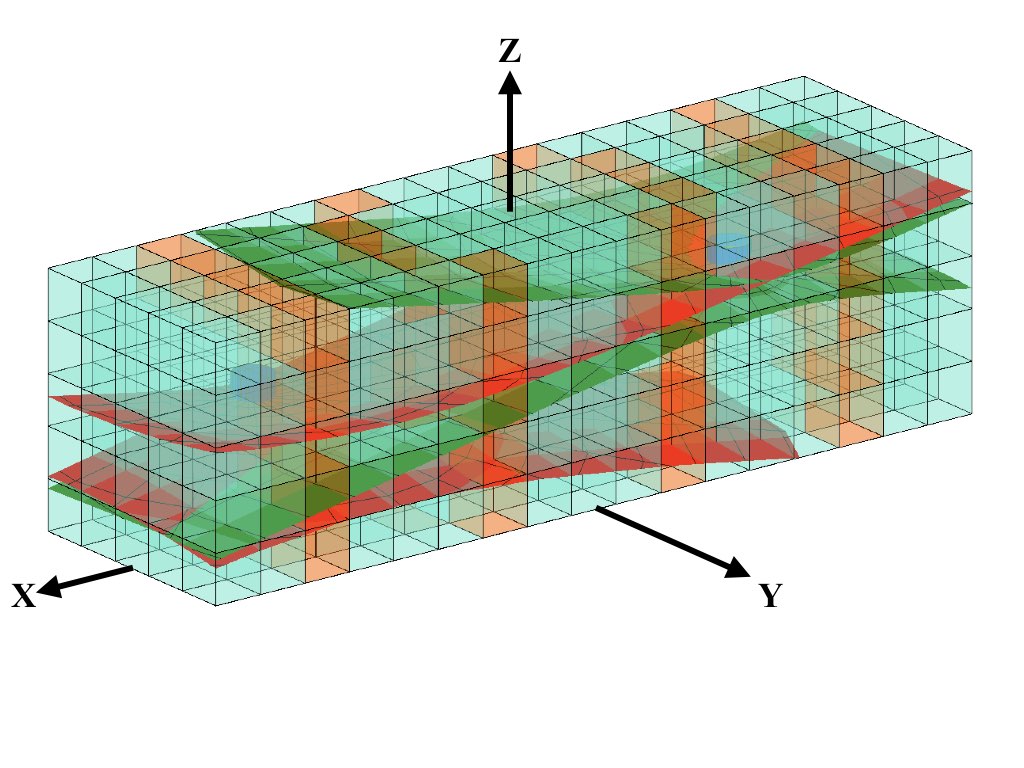}
\caption{Case study with 2 LiDARs, 2 lasers with pitch angles as $\pm 10^\circ$ of each LiDAR.}
\label{case1}	
\end{figure}

In this case, only $[\bm X,\bm Y,\bm Z]$ coordinates are optimized.  The optimization boundary of $[\bm X,\bm Y,\bm Z]$ is the range of ROI. The configuration results of these two LiDARs are: [4.335641 m, -0.777785 m, 0.696529 m] for LiDAR with red laser cone and [-4.335641 m, 1.893497 m, -0.696529 m] for LiDAR with green laser cone. From the result, our optimization model is validated and our method proves to be reliable to find the most informative perception.

\section{Conclusion and Future Work}

In this paper, we proposed a model of the optimal LiDAR configuration for self-driving cars and validated its effectiveness. In the model, the sparsity and discreteness in LiDAR perception were considered in a predefined LiDAR's range of interest (ROI). In order to make the optimal configuration solvable, we proposed a lattice-based model and cylinder-based approximated model. The most informative perception was acquired when the configuration achieves the minimal value of the largest inscribed sphere radius. Finally, we presented a simple case study to demonstrate the effectiveness of our proposed approach.

However, the rotating time duration needed for a laser to form a cone was not considered. In the future, we will discrete the time duration and try to find out the optimal configuration of LiDAR in this perspective. The occlusion between LiDAR will be added to generate a set of subspaces in a more real form. Moreover, the decision variables generated by this method increases exponentially along the number of LiDARs and lasers of each LiDAR, which takes a very huge of computation cost, even for several days. Therefore, a more cost-efficient method for optimization model is needed. Also, Gurobi is not able to solve complex nonlinear programming, which does not allow angles to be solved and thus motivates us to try different solvers in the next works.

\section*{Appendix}
\subsection{Coordinate Transformation}
The transformation of car frame coordinates of a cube to a LIDAR frame coordinates

$$
	R_y(\beta_l) = 
    \begin{bmatrix}
    \cos\beta_l & 0 & \sin\beta_l\\
    0 & 1 & 0\\
    -\sin\beta_l & 0 & \cos\beta_l
    \end{bmatrix}
$$
$$
	R_x(\gamma_l) = 
    \begin{bmatrix}
    1 & 0 & 0\\
    0 & \cos\gamma_l & -\sin\gamma_l\\
    0 & \sin\gamma_l & \cos\gamma_l
    \end{bmatrix}
$$
$$
	\mathbf{R_\textit{l}} = R_y(\beta)R_x(\gamma)
$$
$$
	\mathbf{T_\textit{l}} = [x_l , y_l , z_l]^T
$$
\begin{displaymath}
\mathbf{H_\textit{l}}
=
\begin{bmatrix}
\mathbf{R_\textit{l}} & \mathbf{T_\textit{l}}\\
\mathbf{0} & 1

\end{bmatrix}
\end{displaymath}

\subsection{Logic Constraints}

\subsubsection{\textbf{AND} logic}
Let $\delta_{i}$, $\delta$, $d_{i}$ and $d$ be binary variables
$$
-\sum\limits_{i=1}^{n}\delta_{i} + \delta \leq \epsilon\\
$$
$$
\sum\limits_{i=1}^{n}\delta_{i} - M\delta\leq \epsilon
$$

$M$ is a large positive constant and $\epsilon$ is a small positive constant. In our simulation, M = 200 and $\epsilon$ = 0.01. The value of M and $\epsilon$ may change according to the number of cubes. Let $d_i = 1-\delta_i$ and $d = 1-\delta$, so that 
$$d = AND[d_1,\cdots,d_n]$$

\subsubsection{\textbf{OR} logic}
Let $f_{i}$ and $f$ be binary variables
$$
-\sum\limits_{i=1}^{n}f_{i} + f \leq \epsilon\\
$$
$$
\sum\limits_{i=1}^{n}f_{i} - Mf\leq \epsilon
$$

Here, $M$ and $\epsilon$ are set as in \textbf{AND} logic, therefore, we have
$$f = OR[f_1,\cdots,f_n]$$

\subsubsection{\textbf{IF THEN ELSE} logic\cite{bemporad1999control}}
\textbf{IF} $f(x) \leq 0$ \textbf{THEN} d = 1 \textbf{ELSE} d = 0
$$
f(x)\leq M(1-d)
$$
$$
f(x)\geq \epsilon - (M+\epsilon)d
$$
$M$ and $\epsilon$ are set as same as in $\textbf{OR}$ logic.

\subsection{Definition of variables}
Variable notations in this paper are listed in Table \ref{table:variable}.
\begin{table}[h]
    \caption{Variables Definition}\label{table:variable}
    \begin{tabular}{l p{7cm}} 
     \hline \hline
      Variable & Definition\\
      \hline
      $N_l$ & number of LiDARs \\
      $N_r$ & number of lasers of each LiDAR \\
      $N_s$ & $N_lN_r$: Total number of lasers \\
      $N_c$ & number of cubes \\
      $N_k$ & number of cylinders \\
      $N_{ss}$& $(N_r+1)^{N_l}$: total number of subspaces\\
      $\theta_{lr}$ & vertical angle of $r$th laser on $l$th LiDAR \\
      $f_{lr}$ & binary flag to change the sign of $lr$th inequality\\
      $E_{sck}$ & binary value to determine whether the $c$-th cube of the $k$-th cylinder is in the $s$-th subspace\\
      $x_l$ & $x$ car frame coordinates of $l$th LiDAR\\
      $y_l$ & $y$ car frame coordinates of $l$th LiDAR\\
      $z_l$ &  $z$ car frame coordinates of $l$th LiDAR\\
      $\beta_l$ & Pitch angle of $l$th LiDAR\\
      $\gamma_l$ & Roll angel of $l$th LiDAR\\
      $\bm X$ & Set of $x$ car frame coordinates of all LiDARs\\
      $\bm Y$ & Set of $y$ car frame coordinates of all LiDARs\\
      $\bm Z$ & Set of $z$ car frame coordinates of all LiDARs\\
      $\bm\beta$ & Set of pitch angles of all LiDARs\\
      $\bm\gamma$ & Set of roll angels of all LiDARs\\
      $\mathbf{C}$ & $[\bm X,\bm Y,\bm Z,\bm\alpha,\bm\beta,\bm\gamma]$: configuration parameters of LiDAR \\
      $F_s(\mathbf{C})$ & the number of cubes in subspace$_c$\\ \hline \hline
    \end{tabular}
\end{table}

\section*{Acknowledgment}

The authors would like to thank staff of DENSO R\&D Lab, Ann Arbor. This study is supported by Denso International America, Inc. under master alliance program with University of Michigan.

\bibliographystyle{IEEEtran}
\bibliography{ref.bib}

\end{document}